\documentclass[letterpaper]{article} 
\usepackage[preprint]{aaai2027} 
\usepackage[hyphens]{url} 
\usepackage{graphicx} 
\usepackage{natbib} 
\usepackage{caption} 
\usepackage{booktabs}
\usepackage{amsmath}
\usepackage{amssymb}

\title{AT3D-AD: Anomaly Type-Aware 3D Anomaly Detection via Hierarchical Point-Language Alignment}
\author{
Jingyu Zeng, Haoquan Lu, Can Gao\corresponding
}

\affiliations{
College of Computer Science and Software Engineering and the Guangdong Key Laboratory of Intelligent Information Processing,\\
Shenzhen University, Shenzhen 518060, China\\
Email: \texttt{2005gaocan@163.com}
}

\begin{document}

\maketitle

\begin{abstract}
Detecting and localizing 3D point-cloud defects is essential for industrial inspection.
However, existing methods often suffer from imprecise localization due to
the lack of anomaly supervision and reliance on single-granularity representations.
To address these limitations, we propose Anomaly Type-Aware 3D Anomaly Detection (AT3D-AD),
a unified framework for joint detection, localization, and classification.
Specifically, we first design the Physics-Driven Parametric Anomaly Synthesis (PDPAS) module
employing multiple parametric functions to generate synthetic anomalies,
providing explicit anomaly supervision.
Then, we propose the Hierarchical Global–Local Anomaly Alignment (HiGLA) module
to align global and local representations within the normal and anomalous groups.
Finally, we propose the Semantic–Geometric Anomaly Classification (SGAC) module
to jointly learn localization and classification,
yielding spatially precise and type-discriminative anomaly representations.
Extensive experiments establish new state-of-the-art performance on all four benchmarks.
AT3D-AD achieves Object/Point AUROC scores of 98.1\%/98.9\% on
Anomaly-ShapeNet and 95.0\%/95.2\% on Real3D-AD, while reaching 74.2\% Macro-F1 for
anomaly-type recognition on Real3D-AD.
\end{abstract}

\section{Introduction}
3D anomaly detection (3D AD) aims to identify defective objects and localize
anomalous regions in point clouds. It supports automated product screening and
geometric quality control in industrial inspection~\citep{mvtec3dad,m3dm,btf}.
In practical applications, an inspection system is expected to detect anomalous objects,
localize the defects, and identify their defect categories.
However, defective samples are scarce and expensive to annotate,
while normal samples are abundant.
Hence, most 3D AD methods adopt unsupervised training using only normal
samples\citep{real3dad,anomalyshapenet,r3dad,ye2025Po3adPredictingPoint}.

\begin{figure}[!t]
  \centering
  \includegraphics[width=\columnwidth]{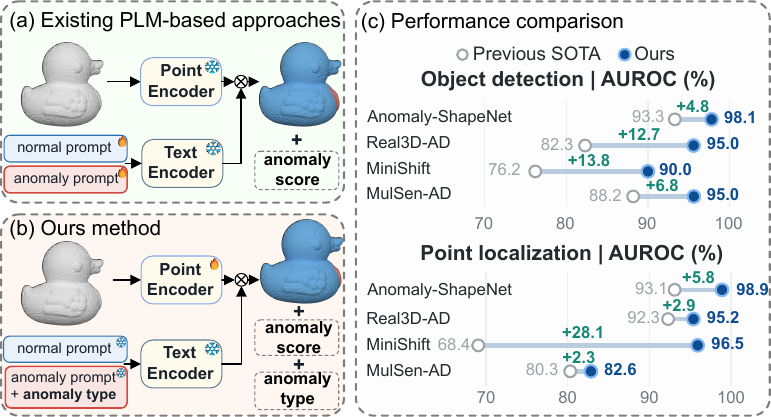}
  \caption{Comparison of existing point-language model (PLM)-based approaches
  and ours. (a) Existing approaches. (b) AT3D-AD. (c) Detection and
  localization performance of AT3D-AD and the previous state of the art on
  four benchmarks. The reported AT3D-AD metrics were obtained using ten-view
  test-time augmentation.}
  \label{fig:teaser}
\end{figure}

Existing methods can be broadly divided into embedding-based and
reconstruction-based methods. Embedding-based methods model defect-free samples
in a feature space and measure test-time deviations from normal features~\citep{btf,m3dm,real3dad}.
Reconstruction-based methods instead learn to recover
normal geometry and identify anomalies from the difference between input and
reconstructed point clouds, as in IMRNet and R3D-AD
\citep{anomalyshapenet,r3dad}.

Despite their effectiveness, these methods face two key limitations.
First, unsupervised training does not provide direct supervision for
the shape or location of the defect.
Hence, localization must be inferred from feature distances or reconstruction errors. 
Although existing methods~\citep{r3dad,ye2025Po3adPredictingPoint}
employ synthetic anomalies for anomaly-aware learning,
they fail to fully exploit the rich structural and semantic information
represented by the generated anomalies.
Second, PointCore combines local and global features,
whereas BTP aligns patch, geometric, and global
representations~\citep{pointcore,li2026BackPointExploring}.
However, they fail to explicitly specify structural anomaly features during training.
Hence, GLFM integrates anomaly synthesis with global-local
feature matching \citep{cheng2025BoostingGlobalLocalFeature}.
Nevertheless, it collapses all defects into a single anomaly class,
overlooking the discriminability of type-specific anomaly semantics.

To address these limitations, this study proposes an
Anomaly Type-Aware 3D Anomaly Detection method (AT3D-AD) for
joint anomaly detection, localization, and type recognition on 3D point clouds.
The differences between existing methods and AT3D-AD are shown in Figure~\ref{fig:teaser}.
AT3D-AD synthesizes bulge, dent, breakage, and scratch defects from normal point clouds,
yielding normal and anomalous labels fed into the hierarchical global-local alignment network.
The network extracts features aligned with shared normal and anomalous semantic prototypes,
enhancing object-level anomaly discrimination and point-level anomaly localization.
Based on the predicted anomalous regions,
AT3D-AD further integrates semantic representations with local geometric features
for fine-grained anomaly-type classification.
By jointly optimizing anomaly localization and type classification,
AT3D-AD enables the semantic and geometric cues learned from different anomaly types
to further refine the localization of anomalous regions.

Our contributions are as follows:
\begin{itemize}
  \item We propose Physics-Driven Parametric Anomaly Synthesis (PDPAS), which
  models four geometric defects in local surface coordinates and
  generates object-level, point-level, and type-level supervision from normal
  point clouds.
  \item We develop Hierarchical Global--Local Anomaly Alignment (HiGLA) to
  align global and local features with shared text prototypes and connect
  object-, region-, and point-level anomaly evidence.
  \item Experiments on four benchmarks show that AT3D-AD achieves
  state-of-the-art detection and localization performance
  together with accurate anomaly-type recognition.
\end{itemize}

\section{Related Work}

\subsection{3D Anomaly Detection}
Unsupervised 3D AD methods broadly follow feature-modeling and reconstruction
paradigms. 
BTF pairs handcrafted Fast Point Feature Histograms (FPFH) with a
PatchCore-style memory bank.
M3DM fuses point and RGB memories, whereas PointCore combines local and global
point features~\citep{rusu2009fpfh,btf,m3dm,pointcore}. The Reg3D-AD baseline
registers test point clouds to normal prototypes before memory-bank retrieval
~\citep{real3dad}. Reconstruction methods instead learn to recover
anomaly-free geometry. IMRNet uses iterative masked reconstruction, R3D-AD uses
diffusion reconstruction, and PO3AD predicts point offsets from pseudo
anomalies~\citep{anomalyshapenet,r3dad,ye2025Po3adPredictingPoint}. 

Synthetic anomaly generation exposes models to diverse abnormal structures and
provides explicit signals for anomaly-aware representation learning.
Anomaly-ShapeNet uses 3D anomaly synthesis to construct diverse defects with
dense annotations~\citep{anomalyshapenet}. R3D-AD uses Patch-Gen to augment
diffusion reconstruction. PO3AD uses normal-vector-guided Norm-AS to supervise
point-offset prediction. GLFM applies point-stretching synthesis before
global-local feature matching
~\citep{r3dad,ye2025Po3adPredictingPoint,cheng2025BoostingGlobalLocalFeature}.

\subsection{Vision-Language Models}
CLIP introduced the image-text contrastive interface used by many zero-shot
recognition and anomaly detection methods~\citep{clip}. In 2D anomaly
detection, WinCLIP and AnomalyCLIP use normal and anomalous text semantics to
produce class-agnostic or few-shot responses~\citep{winclip,anomalyclip}.
PointCLIP and PointCLIP V2 extend this interface to 3D recognition by rendering
point clouds into multiple views~\citep{pointclip,pointclipv2}. 
CLIP3D-AD and PointAD apply a
similar projection-based strategy to few-shot or zero-shot 3D AD
~\citep{clip3dad,pointad}. 

ULIP and ULIP-2 avoid rendering by aligning point clouds, images, and text in a
shared representation space~\citep{ulip,ulip2}.
PLANE adapts ULIP-2 with category-specific static prompts,
sample-specific dynamic prompts, Point Cloud
Feature Adaptation, and Ano3D pseudo anomalies
~\citep{wang2026ExploitingPointlanguageModels}. 
BTP targets zero-shot transfer by integrating
multilayer patch features, global semantics, and FPFH-supervised geometry in
the text-aligned space~\citep{li2026BackPointExploring}. 

However, effective supervision requires discriminative information.
Synthetic supervision methods generate anomalies without type-level supervision.
Conversely, vision-language models provide rich type-level semantic information without
point-level supervision for localization. To address these problems,
we propose an anomaly type-aware 3D AD method,
jointly learning the type-level classification and global-local detection information.

\section{Method}

\begin{figure*}[!ht]
  \centering
  \includegraphics[width=\textwidth]{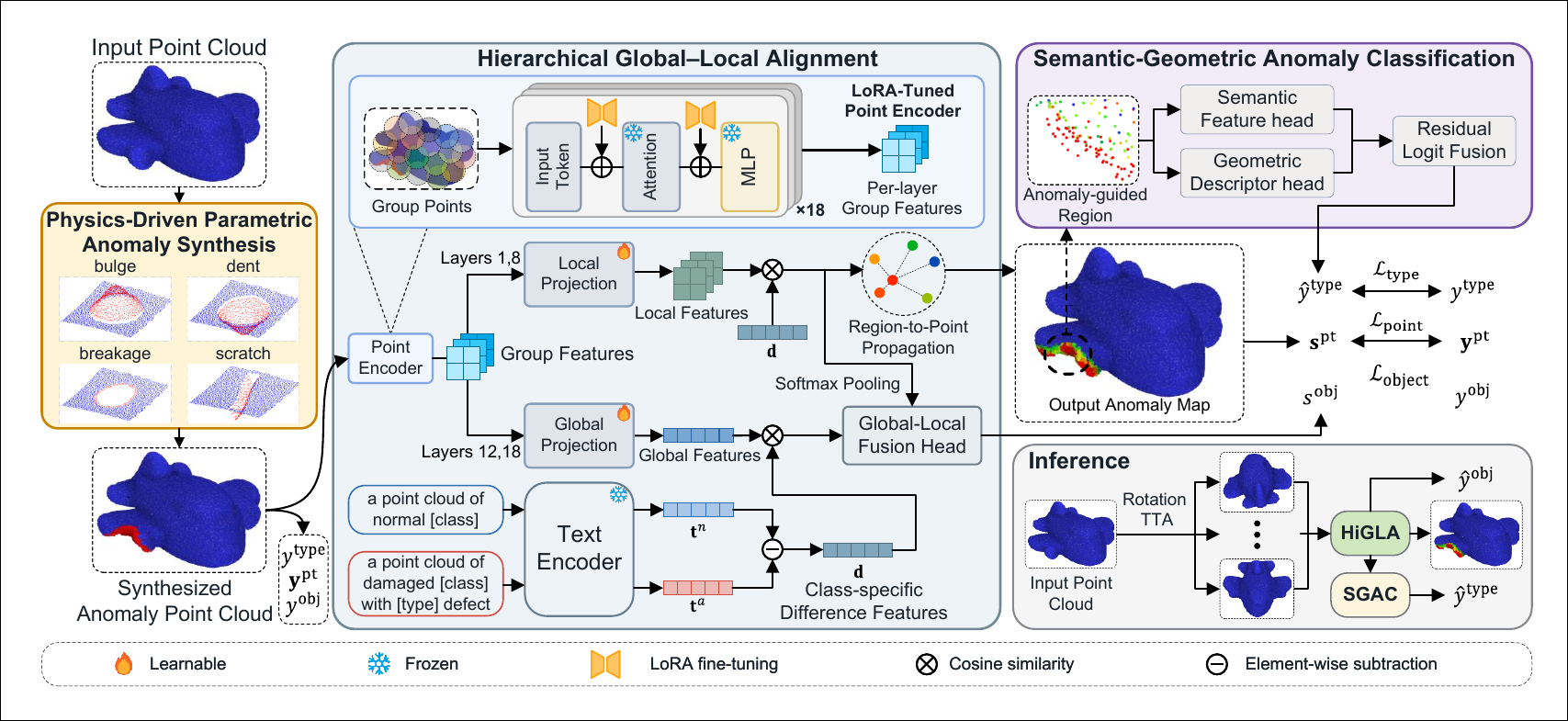}
  \caption{Overview of PDPAS, HiGLA, and SGAC. The complete framework shows
  training supervision, shared modules, information flow, objectives, and
  inference outputs.}
  \label{fig:framework}
\end{figure*}

\subsection{Problem Setting}
Given an input point cloud
$\mathcal{P}=\{\mathbf{p}_i\in\mathbb{R}^{3}\}_{i=1}^{N}$, 3D point-cloud
anomaly detection aims to determine whether $\mathcal{P}$ is anomalous,
localize its anomalous points, and identify its anomaly type. The corresponding
ground-truth labels are defined as
\begin{equation}
y^{\mathrm{obj}}\in\{0,1\},\quad
\mathbf{y}^{\mathrm{pt}}\in\{0,1\}^{N},\quad
y^{\mathrm{type}}\in\{1,\ldots,C\}.
\label{eq:task-labels}
\end{equation}
Here, $y^{\mathrm{obj}}$ and $y_i^{\mathrm{pt}}$ indicate whether the object
and point $i$ are anomalous, respectively. For an anomalous input,
$y^{\mathrm{type}}$ specifies one of $C$ anomaly types.

\subsection{Overview}
AT3D-AD jointly learns object-level detection, point-level localization, and
anomaly-type classification. As shown in Figure~\ref{fig:framework}, it comprises
Physics-Driven Parametric Anomaly Synthesis (PDPAS), Hierarchical Global--Local
Anomaly Alignment (HiGLA), and Semantic--Geometric Anomaly Classification (SGAC).

\subsection{Physics-Driven Parametric Anomaly Synthesis}
Synthetic anomalies provide explicit supervision for anomaly detection.
PDPAS incorporates physical priors associated with three common defect-formation
processes: local deformation, loss of surface support, and elongated surface damage.
For each point cloud, we construct a symmetric $k$-nearest-neighbor ($k$-NN) graph and expand
from a random seed to obtain a connected patch $\Omega$. Local principal
component analysis (PCA) estimates the outward surface normal $\mathbf{n}$,
while projecting a random vector onto the tangent plane provides a tangent
direction, together defining a surface-aligned coordinate frame. The generator
then independently samples the patch size, displacement amplitude, boundary
taper, patch count, and optional deformation cores.

PDPAS applies seven operations to synthesize four canonical anomaly types.
For Bulge and Dent, PDPAS displaces each patch point along the
estimated surface normal:
\begin{equation}
\widetilde{\mathbf{p}}_i
=\mathbf{p}_i+sA w_i\mathbf{n},
\qquad i\in\Omega,
\label{eq:synthesis}
\end{equation}
where $A=\alpha\ell\sqrt{r}>0$ is the scale-adaptive displacement amplitude:
$\alpha$ controls deformation strength, $\ell$ is the median first-neighbor
distance within the patch, and $r$ is the patch-to-cloud size ratio. The spatial
weight $w_i\in[0,1]$ applies a cosine boundary taper, multiplied by a normalized
multi-core Gaussian field when enabled. Finally,
$s\in\{-1,+1\}$ sets the displacement direction, with $s=+1$ for bulges and
$s=-1$ for dents.

Breakage anomalies are synthesized as holes, cracks, or edge chips by
replacing points in a central core, a tangent-aligned band, or a one-sided
peripheral region, respectively, with points resampled from the retained
neighborhood. This operation preserves the input point count. Linear and
curved scratches are generated by applying inward normal displacements
along corresponding surface trajectories.

The seven operations are grouped as Bulge $\{\text{bulge}\}$, Dent
$\{\text{dent}\}$, Breakage $\{\text{hole, crack, edge chip}\}$, and
Scratch $\{\text{scratch, curved scratch}\}$. All patches in a sample share
the same operation and type. Each synthetic sample is assigned
$y^{\mathrm{obj}}=1$, a type label $y^{\mathrm{type}}$, and point-level
labels $\mathbf{y}^{\mathrm{pt}}$ marking the operation-specific affected
region. By integrating these complementary processes into a unified
surface-aligned generator, PDPAS derives synchronized object-, point-, and
type-level supervision entirely from normal data.

\subsection{Hierarchical Global--Local Anomaly Alignment}

Using the synthesized anomaly supervision, HiGLA aligns global and local
point-cloud representations with text-defined anomaly directions.

HiGLA first uses Farthest Point Sampling (FPS) to select $G$ group centers
$\bar{\mathcal P}=\{\bar{\mathbf p}_g\}_{g=1}^{G}$, and each center is
grouped with its $k^{\mathrm{group}}$ nearest neighbors. Then the point
encoder maps each local group to a group token. The group tokens, together with a CLS
token, are then processed by $L$ Transformer layers with LoRA modules inserted into their
attention projections. At layer $\ell$, the Transformer
produces the group features
$\mathbf H^{(\ell)}=[\mathbf h_1^{(\ell)},\ldots,\mathbf h_G^{(\ell)}]^\top\in\mathbb R^{G\times d}$
and the CLS feature $\mathbf h_{\mathrm{cls}}^{(\ell)}\in\mathbb R^d$.

Features from selected Transformer layers are projected and aggregated into
local and global embeddings as follows:
\begin{equation}
\begin{aligned}
\mathbf Z^{\mathrm{loc}}
&=\underset{\ell\in\mathcal I^{\mathrm{loc}}}{\operatorname{Mean}}\left(
\operatorname{Norm}\!\left(\mathbf H^{(\ell)}\mathbf W^{\mathrm{loc}}\right)\right),\\
\mathbf z^{\mathrm{glob}}
&=\underset{\ell\in\mathcal I^{\mathrm{glob}}}{\operatorname{Mean}} \left(
\operatorname{Norm}\!\left(
\mathbf W^{\mathrm{glob}\top}[\mathbf h_{\mathrm{cls}}^{(\ell)}\Vert
\operatorname*{MaxPool}_{g=1}^{G}\mathbf h_g^{(\ell)}]
\right)\right) ,
\end{aligned}
\label{eq:multilevel-aggregation}
\end{equation}
where $\mathbf Z^{\mathrm{loc}}\in\mathbb R^{G\times D}$ and
$\mathbf z^{\mathrm{glob}}\in\mathbb R^D$ denote the aggregated local and
global features, respectively.
The matrices $\mathbf W^{\mathrm{loc}}\in\mathbb R^{d\times D}$ and
$\mathbf W^{\mathrm{glob}}\in\mathbb R^{2d\times D}$ project the local
and global layer features into the shared embedding space.
The operator $\operatorname{Norm}(\cdot)$ denotes
$\ell_2$ normalization along the feature dimension, and $\Vert$ denotes
feature concatenation. The sets $\mathcal I^{\mathrm{loc}}$ and
$\mathcal I^{\mathrm{glob}}$ specify the Transformer layers selected for
local and global aggregation, respectively. 

For an input from object category $o$, the frozen text encoder provides a
normal prototype $\mathbf t_o^n$ and an anomalous prototype
$\mathbf t_{o,c}^a$ for each anomaly type $c$. The corresponding semantic
anomaly direction is
\begin{equation}
\mathbf d_c
=
\frac{\mathbf t_{o,c}^a}{\|\mathbf t_{o,c}^a\|_2}
-
\frac{\mathbf t_o^n}{\|\mathbf t_o^n\|_2},
\qquad c=1,\ldots,C.
\label{eq:semantic-direction}
\end{equation}
Because the object category $o$ is fixed for each input, we omit it from the
notation below.
HiGLA compares the global and local representations with all
type-specific anomaly directions:
\begin{align}
s_c^{\mathrm{glob}}
&=
\tau_{\mathrm{glob}}^{-1}
\left\langle
\mathbf z^{\mathrm{glob}},\mathbf d_c
\right\rangle,\nonumber\\
\mathbf s_c^{\mathrm{loc}}
&=
\tau_{\mathrm{loc}}^{-1}
\left\langle
\mathbf Z^{\mathrm{loc}},\mathbf d_c
\right\rangle.
\label{eq:type-specific-scores}
\end{align}
The local inner product is computed row-wise, yielding
$\mathbf s_c^{\mathrm{loc}}\in\mathbb R^G$.
The resulting scores are then smoothly aggregated over the $C$ anomaly
types to obtain the global score $s^{\mathrm{glob}}$ and the group-level local
scores $\mathbf s^{\mathrm{loc}}\in\mathbb R^G$.

HiGLA propagates type-specific group scores to each point using its $K$ nearest
group centers $\mathcal N_K(i)$:
\begin{align}
w_{ig}&=\underset{g\in\mathcal N_K(i)}{\operatorname{softmax}}
\left(-\|\mathbf p_i-\bar{\mathbf p}_g\|_2/\tau_{\mathrm{dist}}\right),\nonumber\\
s_{i,c}^{\mathrm{pt}}&=\sum_{g\in\mathcal N_K(i)}w_{ig}s_{g,c}^{\mathrm{loc}}.
\label{eq:point-propagation}
\end{align}
Here, $\tau_{\mathrm{dist}}$ controls distance weighting. Applying
$\mathcal A_{\tau_{\mathrm{type}}}$ over the
$C$ anomaly types gives the final point scores
$\mathbf s^{\mathrm{pt}}\in\mathbb R^N$, which are supervised by
$\mathbf y^{\mathrm{pt}}$ with focal, Dice, and ranking losses. A regional KL loss
further aligns the mask-derived target distribution with the predicted group distribution.

HiGLA pools the group-level scores into a local object-level cue
$s^{\mathrm{loc}}=\sum_{g=1}^{G}\rho_gs_g^{\mathrm{loc}}$, where
$\rho_g$ is a temperature-scaled softmax weight. A zero-initialized
local-to-global (L2G) residual head then fuses the detached global and local
scores:
\begin{equation}
s^{\mathrm{obj}}=\bar{s}^{\mathrm{glob}}+
F_{\mathrm{L2G}}\!\left(\bar{s}^{\mathrm{glob}},
\bar{s}^{\mathrm{loc}}\right),
\label{eq:l2g}
\end{equation}
where $\bar{s}=\operatorname{stopgrad}(s)$. Zero initialization makes the
initial output equal to $s^{\mathrm{glob}}$, while gradient detachment ensures
that the fusion loss updates only $F_{\mathrm{L2G}}$.

\subsection{Semantic--Geometric Anomaly Classification}
SGAC predicts the anomaly type from the region localized by HiGLA. During
training, the region mask $\mathbf m$ contains a random subset of anomalous
points with limited normal context; during inference, it is obtained from
the highest-scoring points. The mask induces normalized group weights
$\boldsymbol{\sigma}\in\mathbb R^G$, yielding the region representation
\begin{equation}
\mathbf r
=
\boldsymbol{\sigma}^{\top}\mathbf Z^{\mathrm{loc}}.
\label{eq:region-pooling}
\end{equation}

SGAC combines semantic and geometric type evidence:
\begin{align}
\mathbf u
&=
H_{\mathrm{sem}}
\left(
\mathbf r,\mathbf z^{\mathrm{glob}},\mathbf T_o
\right),\nonumber\\
\mathbf v
&=
H_{\mathrm{geo}}
\left(
\boldsymbol{\phi}(\mathcal P,\mathbf m)
\right),\nonumber\\
\boldsymbol{\ell}^{\mathrm{type}}
&=
\mathbf u+\mathbf v+
H_{\mathrm{fuse}}
\left(
[\mathbf u;\mathbf v;\mathbf u-\mathbf v;
\mathbf u\odot\mathbf v]
\right).
\label{eq:sgac}
\end{align}
Here, $\mathbf T_o\in\mathbb R^{C\times D}$ contains the anomaly-type text
prototypes for object category $o$, and
$\mathbf u,\mathbf v,\boldsymbol{\ell}^{\mathrm{type}}\in\mathbb R^C$.
The composite semantic head $H_{\mathrm{sem}}$ combines an MLP over the region
feature, global context, and their difference with centered prototype-similarity
calibration.
The geometric descriptor $\boldsymbol{\phi}$ combines FPFH
statistics~\citep{rusu2009fpfh}, signed surface profiles, and local
morphology. The geometric and fusion heads are MLPs.

For SGAC, the group tokens and global feature are detached from the
backbone. The type loss therefore updates the SGAC heads and
$\mathbf W^{\mathrm{loc}}$ without altering the PointBERT features.

\subsection{Training Objective and Inference}
The overall objective jointly optimizes object-level detection, point-level
localization, and anomaly-type classification:
\begin{equation}
\mathcal L=\mathcal L_{\mathrm{obj}}+
\mathcal L_{\mathrm{point}}+\mathcal L_{\mathrm{type}}.
\label{eq:objective}
\end{equation}
The object-level term applies weighted binary cross-entropy with logits to
$s^{\mathrm{glob}}$, $s^{\mathrm{loc}}$, and $s^{\mathrm{obj}}$ using
$y^{\mathrm{obj}}$. The point-level term combines focal, Dice, and ranking
losses on $\mathbf s^{\mathrm{pt}}$ against $\mathbf y^{\mathrm{pt}}$ with
regional KL alignment. For samples with valid anomaly-type labels, the type
term applies class-balanced, label-smoothed cross-entropy to
$\boldsymbol{\ell}^{\mathrm{type}}$ using $y^{\mathrm{type}}$, augmented by
hard-negative and auxiliary branch losses.

At inference, we apply random $\mathrm{SO}(3)$ rotations as test-time
augmentation (TTA) to reduce orientation-dependent variation in point-cloud
predictions. Object and point scores are averaged over the original point
cloud and its rotated views, while the corresponding group and global
features are averaged for type classification. The averaged point map
determines the highest-scoring regions, which SGAC classifies using the
averaged features.

\section{Experiments}

\begin{table*}[!ht]
  \centering
  {\scriptsize
  \setlength{\tabcolsep}{0pt}
        \begin{tabular*}{\textwidth}{@{\extracolsep{\fill}}l@{\hspace{3pt}}l*{20}{c}@{}}
    \toprule
    \raisebox{1.4ex}[0pt][0pt]{Method} & \raisebox{1.4ex}[0pt][0pt]{Venue} & \rotatebox{30}{ashtray0} & \rotatebox{30}{bag0} & \rotatebox{30}{bottle0} & \rotatebox{30}{bottle1} & \rotatebox{30}{bottle3} & \rotatebox{30}{bowl0} & \rotatebox{30}{bowl1} & \rotatebox{30}{bowl2} & \rotatebox{30}{bowl3} & \rotatebox{30}{bowl4} & \rotatebox{30}{bowl5} & \rotatebox{30}{bucket0} & \rotatebox{30}{bucket1} & \rotatebox{30}{cap0} & \rotatebox{30}{cap3} & \rotatebox{30}{cap4} & \rotatebox{30}{cap5} & \rotatebox{30}{cup0} & \rotatebox{30}{cup1} & \rotatebox{30}{eraser0} \\
    \midrule
    IMRNet & CVPR24 & 67.1 & 66.0 & 55.2 & 70.0 & 64.0 & 68.1 & 70.2 & 68.5 & 59.9 & 67.6 & 71.0 & 58.0 & 77.1 & 73.7 & 77.5 & 65.2 & 65.2 & 64.3 & 75.7 & 54.8 \\
    GLFM & IEEE TASE25 & 52.8 & 53.7 & 49.5 & 69.9 & 74.5 & 53.1 & 54.3 & 61.0 & 79.4 & 75.1 & 72.0 & 51.2 & 66.2 & 62.1 & 56.4 & 81.2 & 64.2 & 56.4 & 63.4 & 55.7 \\
    PLANE & ESWA25 & 90.5 & 91.4 & 84.3 & 81.4 & \underline{99.4} & 96.3 & 90.7 & 95.6 & 70.6 & 89.3 & 80.0 & 98.1 & \textbf{96.8} & 94.4 & 95.4 & 73.0 & 87.0 & 80.5 & 70.5 & \textbf{100.0} \\
    PO3AD & CVPR25 & \textbf{100.0} & 83.3 & 90.0 & 93.3 & 92.6 & 92.2 & 82.9 & 83.3 & 88.1 & 98.1 & 84.9 & 85.3 & 78.8 & 87.7 & 85.9 & 79.2 & 67.0 & 87.1 & 83.3 & \underline{99.5} \\
    MC3D-AD & IJCAI25 & 96.2 & 80.5 & 79.5 & 70.9 & 75.6 & 93.0 & \textbf{97.8} & 71.9 & 88.5 & 91.1 & 75.4 & 89.8 & 78.4 & 79.3 & 70.1 & 83.5 & 76.1 & 74.3 & 95.2 & 77.6 \\
    Simple3D & AAAI26 & \underline{99.5} & 88.1 & 97.6 & 95.1 & \textbf{100.0} & \textbf{100.0} & 83.0 & 71.1 & 91.1 & 73.0 & 86.3 & 95.9 & 79.0 & 85.2 & 86.7 & 91.2 & 80.4 & \textbf{100.0} & 82.4 & \textbf{100.0} \\
    CASL & AAAI26 & 94.3 & \underline{94.8} & 95.7 & \underline{95.4} & \textbf{100.0} & \textbf{100.0} & 93.3 & \underline{99.3} & \underline{99.6} & 94.4 & 88.8 & \textbf{99.0} & 91.4 & 94.8 & 82.1 & 77.2 & 55.4 & 99.0 & 64.8 & \underline{99.5} \\
    PA3AD & PR26 & \textbf{100.0} & 93.8 & 98.1 & 93.3 & 97.1 & 98.9 & 89.6 & 95.6 & 90.4 & \underline{98.9} & 92.6 & 96.8 & 89.5 & 94.4 & 93.3 & 90.5 & 92.6 & \textbf{100.0} & 95.7 & \textbf{100.0} \\
    SeDiR & CVPR26 & 97.6 & 88.6 & 92.9 & 91.6 & 98.4 & 95.9 & 95.6 & 92.6 & 96.7 & 98.5 & 97.9 & 91.1 & 93.3 & \underline{98.5} & \underline{98.9} & \textbf{99.6} & 87.4 & \underline{99.5} & \textbf{100.0} & 66.7 \\
    MFF-M3AD & NN26 & 89.0 & 80.0 & 85.2 & 76.8 & 94.0 & \underline{99.2} & 92.2 & 86.7 & 93.3 & 84.8 & \underline{98.9} & 84.1 & 90.8 & 89.3 & 81.8 & 97.2 & \textbf{95.4} & 97.6 & 97.6 & 74.3 \\
    Ours & -- & \textbf{100.0} & \textbf{100.0} & \underline{98.6} & \textbf{100.0} & \textbf{100.0} & \textbf{100.0} & 95.6 & \textbf{100.0} & \underline{99.6} & \textbf{100.0} & \textbf{100.0} & 97.1 & 93.3 & 95.6 & 96.1 & \underline{97.9} & 92.3 & \textbf{100.0} & \underline{99.5} & \textbf{100.0} \\
    Ours (TTA=10) & -- & \textbf{100.0} & \textbf{100.0} & \textbf{100.0} & \textbf{100.0} & \textbf{100.0} & \textbf{100.0} & \underline{97.0} & \textbf{100.0} & \textbf{100.0} & \textbf{100.0} & \textbf{100.0} & \underline{98.7} & \underline{94.6} & \textbf{98.9} & \textbf{99.0} & 97.5 & \underline{95.1} & \textbf{100.0} & \underline{99.5} & \textbf{100.0} \\
    \bottomrule
  \end{tabular*}

    \begin{tabular*}{\textwidth}{@{\extracolsep{\fill}}l@{\hspace{3pt}}l*{21}{c}@{}}
    \toprule
    \raisebox{1.4ex}[0pt][0pt]{Method} & \raisebox{1.4ex}[0pt][0pt]{Venue} & \rotatebox{30}{headset0} & \rotatebox{30}{headset1} & \rotatebox{30}{helmet0} & \rotatebox{30}{helmet1} & \rotatebox{30}{helmet2} & \rotatebox{30}{helmet3} & \rotatebox{30}{jar0} & \rotatebox{30}{phone0} & \rotatebox{30}{shelf0} & \rotatebox{30}{tap0} & \rotatebox{30}{tap1} & \rotatebox{30}{vase0} & \rotatebox{30}{vase1} & \rotatebox{30}{vase2} & \rotatebox{30}{vase3} & \rotatebox{30}{vase4} & \rotatebox{30}{vase5} & \rotatebox{30}{vase7} & \rotatebox{30}{vase8} & \rotatebox{30}{vase9} & Mean \\
    \midrule
    IMRNet & CVPR24 & 72.0 & 67.6 & 59.7 & 60.0 & 64.1 & 57.3 & 78.0 & 75.5 & 60.3 & 67.6 & 69.6 & 53.3 & 75.7 & 61.4 & 70.0 & 52.4 & 67.6 & 63.5 & 63.0 & 59.4 & 66.1 \\
    GLFM & IEEE TASE25 & 59.2 & 63.4 & 59.3 & 62.8 & 56.5 & 53.8 & 56.9 & 71.9 & 57.0 & 68.9 & 42.6 & 66.7 & 76.1 & 59.3 & 67.3 & 57.3 & 54.8 & 51.3 & 66.3 & 70.3 & 61.9 \\
    PLANE & ESWA25 & 78.2 & 77.6 & 70.4 & 54.3 & \textbf{100.0} & 72.1 & \textbf{100.0} & \textbf{100.0} & 75.9 & 46.7 & 65.2 & 89.6 & 77.1 & 97.1 & 78.2 & 77.3 & 69.0 & 93.8 & \textbf{96.4} & 59.2 & 83.6 \\
    PO3AD & CVPR25 & 80.8 & 92.3 & 76.2 & 96.1 & 86.9 & 75.4 & 86.6 & 77.6 & 57.3 & 74.5 & 68.1 & 85.8 & 74.2 & 95.2 & 82.1 & 67.5 & 85.2 & \underline{96.6} & 73.9 & 83.0 & 83.9 \\
    MC3D-AD & IJCAI25 & 86.2 & 88.6 & 67.2 & \textbf{100.0} & 60.9 & 97.9 & 97.1 & 91.9 & 84.1 & 94.5 & \underline{97.0} & 82.1 & 85.7 & 92.9 & 76.1 & 87.6 & 97.6 & 93.8 & 67.0 & 73.6 & 84.2 \\
    Simple3D & AAAI26 & \underline{98.2} & 95.7 & 69.0 & 71.9 & 75.4 & 65.5 & 90.5 & \textbf{100.0} & 74.8 & 70.3 & 59.6 & 95.4 & 82.4 & 87.1 & 81.2 & 86.4 & 96.2 & 89.5 & 85.5 & 81.5 & 86.0 \\
    CASL & AAAI26 & 83.6 & \underline{96.7} & 78.3 & 89.0 & 88.1 & 86.1 & \underline{99.5} & \underline{98.1} & 83.2 & 76.1 & 56.3 & 83.8 & 89.0 & \underline{98.1} & \textbf{93.0} & 82.1 & 71.4 & \textbf{100.0} & 93.3 & 81.8 & 88.7 \\
    PA3AD & PR26 & \textbf{100.0} & \textbf{98.1} & 80.3 & \textbf{100.0} & 90.7 & 87.6 & 95.2 & 86.2 & 70.4 & 92.4 & 78.2 & \underline{97.5} & 90.5 & \underline{98.1} & 91.8 & 78.2 & 98.1 & \textbf{100.0} & 93.3 & \textbf{98.4} & 93.3 \\
    SeDiR & CVPR26 & 84.9 & 89.0 & \underline{96.5} & \textbf{100.0} & 87.0 & \textbf{100.0} & \textbf{100.0} & 90.5 & \underline{88.7} & 92.7 & 89.3 & 88.8 & \textbf{97.1} & 97.6 & 78.8 & 92.4 & 94.8 & \textbf{100.0} & 87.3 & 94.2 & 93.3 \\
    MFF-M3AD & NN26 & 85.8 & 89.5 & 92.5 & \textbf{100.0} & 73.3 & \textbf{100.0} & 99.0 & 94.8 & \textbf{89.3} & 95.5 & \textbf{98.1} & 92.9 & 88.1 & 85.2 & 80.6 & 88.2 & \textbf{100.0} & 64.4 & 75.6 & 83.9 & 89.9 \\
    Ours & -- & \textbf{100.0} & 93.8 & 91.3 & \underline{99.1} & \underline{97.4} & \underline{99.4} & \textbf{100.0} & \textbf{100.0} & 84.6 & \underline{97.3} & 87.8 & \textbf{100.0} & \underline{94.8} & \textbf{100.0} & 90.6 & \underline{93.3} & \underline{98.6} & \textbf{100.0} & 90.6 & 90.6 & \underline{96.9} \\
    Ours (TTA=10) & -- & \textbf{100.0} & 93.8 & \textbf{97.7} & \textbf{100.0} & 96.8 & \underline{99.4} & \textbf{100.0} & \textbf{100.0} & 88.4 & \textbf{99.7} & 93.7 & \textbf{100.0} & \textbf{97.1} & \textbf{100.0} & \underline{92.7} & \textbf{95.2} & \underline{98.6} & \textbf{100.0} & \underline{93.9} & \underline{94.9} & \textbf{98.1} \\
    \bottomrule
  \end{tabular*}
  }
  \caption{Per-category Object AUROC on Anomaly-ShapeNet (\%). Bold and underlined
  values denote the best and second-best results, respectively.}
  \label{tab:anomaly-shapenet-object}
\end{table*}

\begin{table*}[!ht]
  \centering
  {\setlength{\tabcolsep}{2.0pt}
  \begin{tabular}{@{}ll*{13}{c}@{}}
    \toprule
    Method & Venue & \rotatebox{45}{airplane} & \rotatebox{45}{candybar} & \rotatebox{45}{car} & \rotatebox{45}{chicken} & \rotatebox{45}{diamond} & \rotatebox{45}{duck} & \rotatebox{45}{fish} & \rotatebox{45}{gemstone} & \rotatebox{45}{seahorse} & \rotatebox{45}{shell} & \rotatebox{45}{starfish} & \rotatebox{45}{toffees} & Mean \\
    \midrule
    GLFM & IEEE TASE25 & 54.6 & 71.5 & 84.2 & 68.8 & 71.2 & \textbf{94.5} & 69.5 & 68.8 & 92.4 & 73.3 & 74.8 & 76.3 & 75.0 \\
    PLANE & ESWA25 & 63.0 & 77.4 & 83.3 & 69.3 & \underline{99.8} & 83.9 & 93.9 & 89.1 & 54.0 & 86.6 & 60.1 & 71.6 & 77.7 \\
    PO3AD & CVPR25 & 80.4 & 78.5 & 65.4 & 68.6 & 80.1 & 82.0 & 85.9 & 69.3 & 75.6 & 80.0 & 75.8 & 77.1 & 76.5 \\
    MC3D-AD & IJCAI25 & 85.0 & 83.0 & 74.9 & 71.5 & 95.5 & 83.1 & 86.5 & 56.0 & 71.6 & 80.3 & 76.6 & 73.8 & 78.2 \\
    Simple3D & AAAI26 & 76.5 & 65.1 & \textbf{98.1} & 82.6 & \textbf{100.0} & 77.8 & 91.2 & 70.4 & 93.0 & 51.4 & 69.6 & 88.8 & 80.4 \\
    CASL & AAAI26 & 80.8 & \underline{84.8} & 79.9 & 65.7 & 97.6 & 83.6 & 93.5 & 76.9 & 64.3 & 79.1 & 89.3 & 92.4 & 82.3 \\
    PA3AD & PR26 & 76.0 & 76.9 & 75.2 & 67.7 & 77.6 & 78.2 & 95.4 & 72.8 & 87.5 & 80.8 & 79.6 & 78.9 & 78.9 \\
    SeDiR & CVPR26 & 86.0 & 81.9 & 78.3 & 72.9 & 94.8 & \underline{86.2} & 93.8 & 62.7 & 67.4 & 77.9 & 85.4 & 84.5 & 81.0 \\
    MFF-M3AD & NN26 & \underline{90.2} & 77.0 & 85.1 & 74.5 & 98.4 & 84.9 & 94.3 & 59.8 & 76.9 & 74.5 & 80.0 & 78.0 & 81.1 \\
    Ours & -- & 88.6 & \textbf{100.0} & 80.0 & \underline{88.0} & \textbf{100.0} & 77.4 & \underline{97.4} & \underline{92.2} & \textbf{99.6} & \underline{90.8} & \underline{92.4} & \underline{97.0} & \underline{91.9} \\
    Ours (TTA=10) & -- & \textbf{93.3} & \textbf{100.0} & \underline{88.0} & \textbf{92.4} & \textbf{100.0} & 84.0 & \textbf{99.0} & \textbf{95.2} & \underline{98.9} & \textbf{96.1} & \textbf{94.9} & \textbf{98.4} & \textbf{95.0} \\
    \bottomrule
  \end{tabular}
  }
  \caption{Per-category Object AUROC on Real3D-AD (\%). Bold and underlined
  values denote the best and second-best results, respectively.}
  \label{tab:real3dad-object}
\end{table*}

\begin{table*}[!ht]
  \centering
  {\scriptsize
  \setlength{\tabcolsep}{0pt}
        \begin{tabular*}{\textwidth}{@{\extracolsep{\fill}}l@{\hspace{3pt}}l*{20}{c}@{}}
    \toprule
    \raisebox{1.4ex}[0pt][0pt]{Method} & \raisebox{1.4ex}[0pt][0pt]{Venue} & \rotatebox{30}{ashtray0} & \rotatebox{30}{bag0} & \rotatebox{30}{bottle0} & \rotatebox{30}{bottle1} & \rotatebox{30}{bottle3} & \rotatebox{30}{bowl0} & \rotatebox{30}{bowl1} & \rotatebox{30}{bowl2} & \rotatebox{30}{bowl3} & \rotatebox{30}{bowl4} & \rotatebox{30}{bowl5} & \rotatebox{30}{bucket0} & \rotatebox{30}{bucket1} & \rotatebox{30}{cap0} & \rotatebox{30}{cap3} & \rotatebox{30}{cap4} & \rotatebox{30}{cap5} & \rotatebox{30}{cup0} & \rotatebox{30}{cup1} & \rotatebox{30}{eraser0} \\
    \midrule
    IMRNet & CVPR24 & 67.1 & 66.8 & 55.6 & 70.2 & 64.1 & 78.1 & 70.5 & 68.4 & 59.9 & 57.6 & 71.5 & 58.5 & 77.4 & 71.5 & 70.6 & 75.3 & 74.2 & 64.3 & 68.8 & 54.8 \\
    GLFM & IEEE TASE25 & 74.1 & 75.4 & 80.9 & 70.7 & 83.3 & 83.0 & 62.0 & 73.8 & 87.7 & 59.7 & 58.2 & 60.9 & 69.4 & 97.6 & 89.1 & 92.6 & 87.2 & 73.1 & 55.9 & 60.9 \\
    PLANE & ESWA25 & 73.4 & 90.4 & 83.3 & 90.0 & \underline{94.9} & 81.0 & 87.4 & 83.4 & 88.0 & 85.3 & 79.3 & 85.5 & \underline{94.7} & 90.4 & 95.1 & 89.4 & 84.4 & 83.4 & 84.7 & 96.7 \\
    PO3AD & CVPR25 & 96.2 & 94.9 & 91.2 & 84.4 & 88.0 & 97.8 & 91.4 & 91.8 & 93.5 & 96.7 & 94.1 & 75.5 & 89.9 & 95.7 & 94.8 & 94.0 & 86.4 & 90.9 & 93.2 & 97.4 \\
    MC3D-AD & IJCAI25 & 80.7 & 85.7 & 90.2 & 86.7 & 90.2 & 77.5 & 56.2 & 59.7 & 77.9 & 67.0 & 56.2 & 90.2 & 86.8 & 85.4 & 90.3 & 85.8 & 88.2 & 76.3 & 69.4 & 82.0 \\
    Simple3D & AAAI26 & 92.0 & 95.4 & 97.4 & 72.8 & 83.8 & \underline{98.8} & 95.1 & 93.3 & 99.3 & 92.9 & 97.9 & 72.5 & 92.1 & 98.8 & 96.4 & 97.9 & \underline{96.4} & 97.9 & 93.7 & 97.0 \\
    CASL & AAAI26 & 88.7 & 96.8 & 85.3 & 86.2 & 91.4 & 95.8 & 97.5 & 98.2 & \underline{99.8} & \underline{99.6} & 97.4 & 79.6 & 92.7 & 97.4 & 97.1 & 94.9 & 66.9 & 98.0 & 79.7 & 87.6 \\
    PA3AD & PR26 & 96.3 & 96.5 & 95.8 & 91.5 & 94.0 & 97.1 & 88.3 & 93.4 & 96.4 & 98.4 & 94.8 & 90.0 & 92.0 & 95.9 & 95.5 & 94.5 & 96.3 & 97.5 & 95.8 & 98.0 \\
    SeDiR & CVPR26 & 76.4 & 89.8 & 91.1 & 89.8 & 94.5 & 86.6 & 70.0 & 79.5 & 86.9 & 80.6 & 64.6 & 71.7 & 88.4 & 90.0 & \underline{97.2} & 94.0 & 95.0 & 87.1 & 84.2 & 74.8 \\
    MFF-M3AD & NN26 & 79.1 & 84.8 & 92.2 & 87.5 & 93.0 & 83.3 & 56.1 & 73.4 & 82.8 & 73.1 & 56.8 & 72.1 & 86.3 & 88.1 & 90.8 & 91.0 & 91.9 & 84.9 & 74.8 & 82.3 \\
    Ours & -- & \underline{98.4} & \underline{99.5} & \underline{99.6} & \underline{98.5} & \textbf{99.7} & 98.6 & \underline{98.5} & \underline{99.8} & \textbf{99.9} & \textbf{99.9} & \underline{99.4} & \underline{98.2} & \textbf{98.6} & \underline{99.3} & \textbf{99.9} & \underline{99.8} & \textbf{99.6} & \underline{99.4} & \underline{98.1} & \underline{99.6} \\
    Ours (TTA=10) & -- & \textbf{98.6} & \textbf{99.7} & \textbf{99.7} & \textbf{98.7} & \textbf{99.7} & \textbf{99.3} & \textbf{98.9} & \textbf{99.9} & \textbf{99.9} & \textbf{99.9} & \textbf{99.7} & \textbf{98.5} & \textbf{98.6} & \textbf{99.4} & \textbf{99.9} & \textbf{99.9} & \textbf{99.6} & \textbf{99.6} & \textbf{98.6} & \textbf{99.7} \\
    \bottomrule
  \end{tabular*}

    \begin{tabular*}{\textwidth}{@{\extracolsep{\fill}}l@{\hspace{3pt}}l*{21}{c}@{}}
    \toprule
    \raisebox{1.4ex}[0pt][0pt]{Method} & \raisebox{1.4ex}[0pt][0pt]{Venue} & \rotatebox{30}{headset0} & \rotatebox{30}{headset1} & \rotatebox{30}{helmet0} & \rotatebox{30}{helmet1} & \rotatebox{30}{helmet2} & \rotatebox{30}{helmet3} & \rotatebox{30}{jar0} & \rotatebox{30}{phone0} & \rotatebox{30}{shelf0} & \rotatebox{30}{tap0} & \rotatebox{30}{tap1} & \rotatebox{30}{vase0} & \rotatebox{30}{vase1} & \rotatebox{30}{vase2} & \rotatebox{30}{vase3} & \rotatebox{30}{vase4} & \rotatebox{30}{vase5} & \rotatebox{30}{vase7} & \rotatebox{30}{vase8} & \rotatebox{30}{vase9} & Mean \\
    \midrule
    IMRNet & CVPR24 & 70.5 & 47.6 & 59.8 & 60.4 & 64.4 & 66.3 & 76.5 & 74.2 & 60.5 & 68.1 & 69.9 & 53.5 & 68.5 & 61.4 & 40.1 & 52.4 & 68.2 & 59.3 & 63.5 & 69.1 & 65.0 \\
    GLFM & IEEE TASE25 & 72.6 & 60.8 & 75.3 & 62.6 & 81.8 & 72.8 & 74.9 & 81.4 & 64.4 & 70.6 & 78.3 & 77.8 & 93.4 & 61.4 & 76.7 & 70.6 & 61.5 & 75.8 & 95.2 & 74.4 & 74.5 \\
    PLANE & ESWA25 & 68.9 & 70.4 & 77.9 & 62.1 & 92.3 & 79.2 & 91.1 & 92.1 & 88.1 & 44.1 & 47.4 & 84.7 & 65.6 & 82.3 & 78.5 & 82.6 & 72.7 & 92.6 & 95.9 & 76.6 & 82.1 \\
    PO3AD & CVPR25 & 82.3 & 90.7 & 87.8 & 94.8 & 93.2 & 84.6 & 87.1 & 81.0 & 66.3 & 78.3 & 69.2 & 95.5 & 88.2 & 97.8 & 88.4 & 90.2 & 93.7 & 98.2 & 95.0 & 95.2 & 89.8 \\
    MC3D-AD & IJCAI25 & 66.6 & 59.2 & 74.9 & 59.1 & 81.8 & 58.5 & 84.7 & 89.1 & 62.5 & 50.2 & 58.4 & 89.7 & 60.8 & 78.1 & 80.0 & 77.2 & 58.8 & 57.6 & 87.4 & 76.2 & 74.8 \\
    Simple3D & AAAI26 & 93.6 & \underline{95.8} & 90.2 & 90.2 & 94.7 & 92.8 & 97.8 & 96.4 & 90.1 & 85.7 & 81.1 & 93.4 & 80.7 & 98.6 & 91.0 & 98.5 & 96.6 & 99.0 & 97.5 & 91.3 & 92.9 \\
    CASL & AAAI26 & 77.0 & 90.9 & 92.3 & 88.2 & 92.9 & 98.0 & 97.4 & 83.8 & 92.8 & 65.2 & 60.2 & 89.4 & \textbf{97.2} & 97.1 & 93.1 & 91.3 & 73.2 & \underline{99.7} & 98.9 & 88.6 & 89.9 \\
    PA3AD & PR26 & 86.1 & 92.2 & 84.6 & 96.1 & 95.2 & 89.9 & 95.3 & 87.9 & 78.0 & 89.4 & 81.7 & 97.4 & 93.9 & 98.3 & 88.9 & 89.1 & 96.2 & 99.2 & 95.0 & 92.0 & 93.1 \\
    SeDiR & CVPR26 & 70.8 & 68.7 & 81.0 & 59.6 & 88.5 & 70.1 & 91.1 & 89.1 & 67.9 & 59.7 & 63.0 & 84.6 & 72.2 & 85.0 & 83.0 & 88.4 & 76.6 & 77.7 & 90.0 & 83.5 & 81.0 \\
    MFF-M3AD & NN26 & 67.1 & 68.9 & 76.1 & 54.7 & 82.3 & 68.2 & 87.0 & 87.6 & 62.8 & 79.2 & 63.8 & 85.5 & 64.0 & 78.2 & 84.8 & 79.2 & 56.4 & 73.5 & 89.7 & 74.6 & 77.1 \\
    Ours & -- & \underline{97.8} & \textbf{97.2} & \underline{97.5} & \underline{98.7} & \underline{97.7} & \underline{98.9} & \underline{99.6} & \underline{98.2} & \underline{95.2} & \underline{98.8} & \underline{91.8} & \underline{98.9} & 93.8 & \underline{99.8} & \underline{98.2} & \underline{99.5} & \underline{99.3} & \textbf{99.9} & \underline{99.5} & \textbf{98.1} & \underline{98.6} \\
    Ours (TTA=10) & -- & \textbf{98.2} & \textbf{97.2} & \textbf{97.9} & \textbf{99.3} & \textbf{98.1} & \textbf{99.4} & \textbf{99.8} & \textbf{98.9} & \textbf{97.2} & \textbf{98.9} & \textbf{94.3} & \textbf{99.2} & \underline{94.0} & \textbf{99.9} & \textbf{98.8} & \textbf{99.8} & \textbf{99.5} & \textbf{99.9} & \textbf{99.6} & \underline{97.9} & \textbf{98.9} \\
    \bottomrule
  \end{tabular*}
  }
  \caption{Per-category Point AUROC on Anomaly-ShapeNet (\%). Bold and underlined
  values denote the best and second-best results, respectively.}
  \label{tab:anomaly-shapenet-point}
\end{table*}

\subsection{Experimental Setup}

\paragraph{Datasets and Metrics.}
We evaluate AT3D-AD on Anomaly-ShapeNet~\citep{anomalyshapenet},
Real3D-AD~\citep{real3dad}, MiniShift~\citep{minishift}, and
MulSen-AD~\citep{mulsenad}. Type evaluation includes all four types on
Anomaly-ShapeNet. On Real3D-AD, it uses the available Bulge and Dent labels.
MiniShift and MulSen-AD evaluate only detection and localization because they
do not provide defect-type labels.

We report object-level AUROC, point-level AUROC and AUPRO following
previous anomaly-detection protocols~\citep{real3dad},
macro-F1 for anomaly-type classification, and both single-view and
$N_{\mathrm{TTA}}=10$ rotation-ensemble results.

\paragraph{Implementation details.}
We initialize the network from the released ULIP-2 PointBERT checkpoint
\citep{ulip2,pointbert}.
Inputs are downsampled to 10,000 points, except for MiniShift,
where we sample 20,000 points.
We train a separate model for each dataset for 250 epochs
and evaluate its final-epoch checkpoint.
All experiments are conducted on an NVIDIA RTX 3090 GPU.

\paragraph{Compared methods.}
For detection and localization, we compare with IMRNet~\citep{anomalyshapenet},
Reg3D-AD~\citep{real3dad}, R3D-AD~\citep{r3dad},
GLFM~\citep{cheng2025BoostingGlobalLocalFeature},
PLANE~\citep{wang2026ExploitingPointlanguageModels},
PO3AD~\citep{ye2025Po3adPredictingPoint}, MC3D-AD~\citep{mc3dad},
MFF-M3AD~\citep{liang2026MFFM3ADUnifiedReconstruction},
Simple3D~\citep{minishift},
CASL~\citep{zha2026CaslCurvatureaugmentedSelfsupervised},
PA3AD~\citep{pa3ad}, and SeDiR~\citep{kim2026SemanticallyDisentangledUnified}.
For anomaly classification, we compare with an MLP,
PointNet~\citep{qi2017PointNetDeepLearning},
PointNet++~\citep{qi2017PointNetDeepHierarchical},
DGCNN~\citep{wang2019DynamicGraphCNN}, and semantic-only and geometry-only
variants of our classifier.

\subsection{Comparison with State-of-the-Art Methods}

\subsubsection{Anomaly-ShapeNet Results}
Tables~\ref{tab:anomaly-shapenet-object}
and~\ref{tab:anomaly-shapenet-point} report Object AUROC and Point AUROC,
respectively, across all 40 categories. AT3D-AD achieved 96.9/98.6
Object/Point AUROC without TTA and 98.1/98.9 with ten views.
The TTA variant ranked first in 25/38
object/point categories and second in another 10/2. Its point
score ranked among the top two in every category.

\subsubsection{Real3D-AD Results}
Table~\ref {tab:real3dad-object} reports the two
metrics separately. The single-view model achieved 91.9 Object
AUROC, compared with the previous best score of 82.3.
TTA results increased the means to 95.0. It ranked first in nine object
categories and second in two. 
The margin over previous works
is 12.7\% for Object AUROC.

\subsubsection{MiniShift and MulSen-AD Results}
Table~\ref{tab:minishift-mulsenad} combines the macro results on the two datasets.
On MiniShift, the single-view model achieved 88.7/95.5 Object/Point AUROC and
exceeded CASL by 12.5/27.1 points. TTA result increased the scores to
90.0/96.5. On MulSen-AD, the corresponding scores increased from 93.0/81.9 to
95.0/82.6. The TTA model exceeded Simple3D by 6.8/2.3 points and MC3D-AD by
17.7/8.4 points.

\begin{table}[!ht]
  \centering
  \small
  \begin{tabular}{@{}lcccc@{}}
    \toprule
    & \multicolumn{2}{c}{\textit{MiniShift}}
    & \multicolumn{2}{c}{\textit{MulSen-AD}} \\
    \cmidrule(lr){2-3}
    \cmidrule(lr){4-5}
    Method
    & \shortstack{Object\\AUROC}
    & \shortstack{Point\\AUROC}
    & \shortstack{Object\\AUROC}
    & \shortstack{Point\\AUROC} \\
    \midrule
    R3D-AD
    & 68.7 & 54.9
    & 80.6 & 56.6 \\
    Reg3D-AD
    & 51.6 & 52.0
    & 77.0 & 60.3 \\
    GLFM
    & 55.8 & 58.7
    & 78.5 & 66.5 \\
    MC3D-AD
    & 60.2 & 65.2
    & 77.3 & 74.2 \\
    Simple3D
    & 68.6 & 66.2
    & 88.2 & 80.3 \\
    CASL
    & 76.2 & 68.4
    & 84.6 & 77.9 \\
    Ours
    & \underline{88.7} & \underline{95.5}
    & \underline{93.0} & \underline{81.9} \\
    Ours (TTA=10)
    & \textbf{90.0} & \textbf{96.5}
    & \textbf{95.0} & \textbf{82.6} \\
    \bottomrule
  \end{tabular}
  \caption{Macro Object AUROC and Point AUROC on MiniShift and MulSen-AD
  (\%). Bold and underlined values denote the best and second-best results,
  respectively.}
  \label{tab:minishift-mulsenad}
\end{table}

\begin{table}[!ht]
  \centering
  \small
  \setlength{\tabcolsep}{2.4pt}
  \begin{tabular}{@{}lcccc@{}}
    \toprule
    Method & \multicolumn{2}{c}{Anomaly-ShapeNet} & \multicolumn{2}{c}{Real3D-AD} \\
    & Accuracy & Macro-F1 & Accuracy & Macro-F1 \\
    \midrule
    MLP & 51.3 & 47.3 & 67.5 & 63.3 \\
    PointNet & 50.1 & 46.6 & 68.1 & 64.9 \\
    PointNet++ & 50.1 & 46.2 & 64.2 & 61.0 \\
    DGCNN & 50.6 & 47.6 & 64.9 & 61.7 \\
    Semantic-only & 59.7 & \underline{57.3} & 70.8 & 66.6 \\
    Geometry-only & \underline{60.4} & 51.4 & \underline{74.9} & \underline{71.9} \\
    \addlinespace
    Ours & \textbf{74.4} & \textbf{67.6} & \textbf{77.5} & \textbf{74.2} \\
    \bottomrule
  \end{tabular}
  \caption{End-to-end anomaly-type classification. Accuracy and Macro-F1 are
  percentages. Bold and underlined values denote the best and second-best
  results, respectively.}
  \label{tab:type-classification}
\end{table}

\subsection{Anomaly Classification}
Table~\ref{tab:type-classification} compares the complete semantic--geometric
classifier with semantic-only, geometry-only, and standard point-cloud heads
based on PointNet, PointNet++, and DGCNN.
Every classifier receives the same anomaly-region input.
On Real3D-AD, the complete head achieved 77.5\% accuracy and 74.2\% Macro-F1.
It exceeded the stronger single branch by 2.6/2.3 points and the strongest
standard head by 9.4/9.3 points.
On Anomaly-ShapeNet, the complete head achieved 74.4\% accuracy and 67.6\%
Macro-F1. These scores exceeded the stronger single branch by 14.0/10.3 points.

\begin{table}[!ht]
  \centering
  \small
  \setlength{\tabcolsep}{2.2pt}
  \begin{tabular}{@{}lccccccc@{}}
    \toprule
    Synth. & G & L & L2G & SGAC & O-AUROC & P-AUROC & P-AUPRO \\
    \midrule
    None  & -- & -- & -- & -- & 65.96 & 71.11 & 38.75 \\
    PO3AD & \checkmark & \checkmark & \checkmark & --
          & 72.74 & 88.67 & 68.86 \\
    PDPAS & \checkmark & -- & -- & -- & 87.85 & 85.83 & 66.22 \\
    PDPAS & \checkmark & \checkmark & -- & -- & 89.77 & 92.84 & 85.74 \\
    PDPAS & \checkmark & \checkmark & \checkmark & -- & 91.04 & 92.84 & 85.74 \\
    PDPAS & \checkmark & \checkmark & \checkmark & \checkmark
          & \textbf{91.94} & \textbf{94.07} & \textbf{87.21} \\
    \bottomrule
  \end{tabular}
  \caption{Nested module ablation on Real3D-AD (\%). G denotes global
  alignment, and L denotes local alignment. Every trained row uses its
  final-epoch checkpoint.}
  \label{tab:ablation}
\end{table}

\begin{figure}[!ht]
  \centering
  \includegraphics[width=0.6\columnwidth]{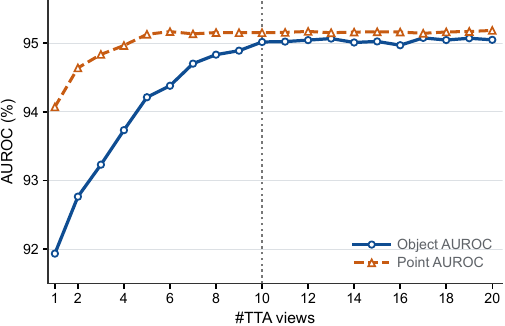}
  \caption{TTA sensitivity on Real3D-AD.
  All other settings are held fixed relative to the single-view evaluation.
  The vertical dotted line marks the ten-view setting used in the main comparison.}
  \label{fig:tta-sensitivity}
\end{figure}

\begin{figure}[!ht]
  \centering
  \includegraphics[width=\columnwidth]{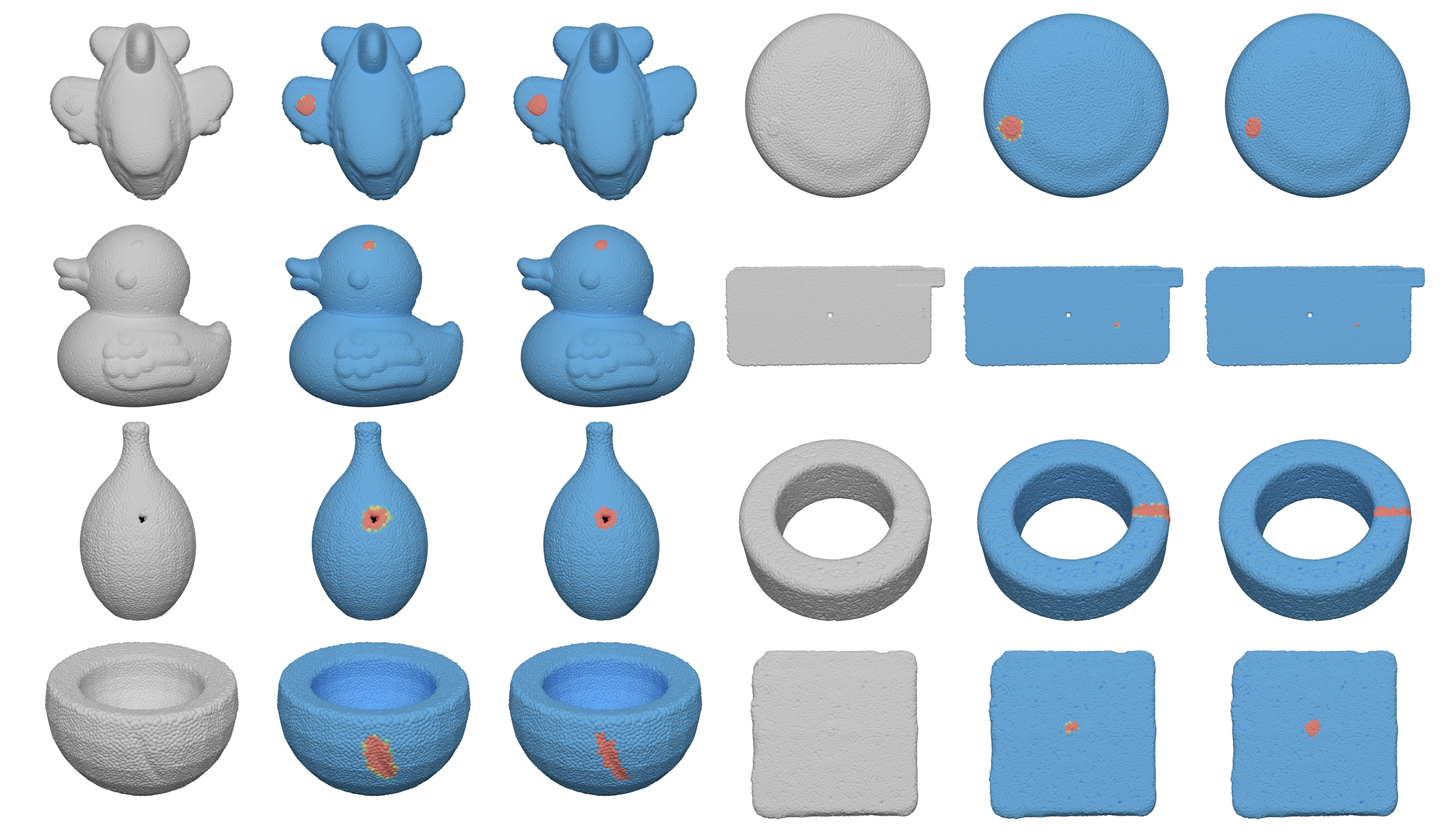}
  \caption{Qualitative comparison shows the input point cloud, AT3D-AD, and the ground truth.}
  \label{fig:qualitative}
\end{figure}

\subsection{Ablation Studies}
Table~\ref{tab:ablation} follows a strictly nested PDPAS chain: global
alignment, local alignment, local-to-global residual fusion, and SGAC are added
one at a time. The PO3AD row is a separate matched-synthesis control. It uses
the same global, local, and L2G modules as the corresponding PDPAS row without
SGAC.

After excluding the matched PO3AD control, the nested PDPAS chain improved
monotonically across all three primary metrics.
Local alignment contributed 7.01 Point-AUROC and 19.52
Point-AUPRO points over the global-only row. L2G affects only the object
readout. It increased Object AUROC from 89.77 to 91.04 without changing the
point map. SGAC further increased the three metrics to 91.94/94.07/87.21.
Under matched G+L+L2G settings, PDPAS exceeded PO3AD by
19.2/5.4/18.35 points.

\paragraph{TTA Sensitivity.}
We first varied the total number of $\mathrm{SO}(3)$ test-time views from 1 to
20 while reusing the same final-epoch checkpoint. On Real3D-AD,
Figure~\ref{fig:tta-sensitivity} shows that both Object and Point AUROC improved
rapidly with additional views before reaching a narrow plateau. We therefore
use ten views in all main comparisons as a practical trade-off between
performance gains and additional inference cost.

\subsection{Qualitative Visualization and Efficiency}
Figure~\ref{fig:qualitative} arranges each example as the raw point cloud, 
the AT3D-AD anomaly map, and the ground-truth.

\paragraph{Efficiency.}
We measured single-view inference on the RTX 3090 using the
Real3D-AD dataset and 10,000-point inputs.
The detection/localization path required 731.29M total parameters,
4.13M trainable parameters, and a median 39.92~ms over
100 runs. Adding SGAC increased these values to 733.43M, 6.27M, and 63.17~ms,
respectively. Thus, SGAC added 2.14M parameters (0.29\% of the full model) and
23.25~ms (58.2\%) per sample.
The SGAC overhead is incurred only when type predictions are
requested. Total counts include the frozen ULIP-2 text encoder.

\section{Conclusion}
This study proposes an Anomaly Type-aware 3D Anomaly Detection (AT3D-AD) method
for industrail inspection. By leveraging synthetic anomaly samples and
semantics-driven anomaly classification, the proposed AT3D-AD addresses
the limitations of previous methods in exploiting anomaly supervision and
multi-granularity semantic information, thereby enabling fine-grained anomaly
localization and detection. The designed Physics-Driven Parametric Anomaly
Synthesis (PDPAS) module generates anomalous samples in four anomaly types,
which enhance the discriminability of model. The Hierarchical Global-Local
Anomaly Alignment (HiGLA) module effectively captures hierarchical global
and local information which are aligned with semantic representations,
facilitating the training of anomaly detection model. Furthermore, the Semantic-Geometric
Anomaly Classification (SGAC) module capitalizes on the jointly learning anomaly
detection and type classification, enabling precise anomaly localization.
Extensive experiments demonstrate the superiority of AT3D-AD, achieving
state-of-the-art performance on four benchmark datasets for 3D AD.
Future work will focus on improving the model's generalization ability on
zero-shot settings.

\bibliography{references}

\end{document}